\documentclass{article}

\usepackage[preprint]{nips_2018}

\usepackage[utf8]{inputenc} % allow utf-8 input
\usepackage[T1]{fontenc}    % use 8-bit T1 fonts
\usepackage{hyperref}       % hyperlinks
\usepackage{url}            % simple URL typesetting
\usepackage{booktabs}       % professional-quality tables
\usepackage{amsfonts}       % blackboard math symbols
\usepackage{nicefrac}       % compact symbols for 1/2, etc.
\usepackage{microtype}      % microtypography

\usepackage{graphicx}
\graphicspath{ {Pictures/} }
\usepackage[labelformat=simple]{subcaption}

\usepackage{float}
\floatstyle{plaintop}
\restylefloat{table}

\PassOptionsToPackage{numbers}{natbib}

\title{Learning End-to-end Autonomous Driving using Guided Auxiliary Supervision}

\author{
  Ashish Mehta \\
  Intel Corporation\\
  \texttt{ashish.mehta@intel.com} \\
   \And
   Adithya Subramanian \\
   Intel Corporation \\
   \texttt{adithya.subramanian@intel.com} \\
   \And
   Anbumani Subramanian \\
   Intel Corporation \\
   \texttt{anbumani.subramanian@intel.com} \\
}

\begin{document}

\maketitle

\begin{abstract}
 Learning to drive faithfully in highly stochastic urban settings remains an open problem. To that end, we propose a Multi-task Learning from Demonstration (MT-LfD) framework which uses supervised auxiliary task prediction to guide the main task of predicting the driving commands. Our framework involves an end-to-end trainable network for imitating the expert demonstrator's driving commands. The network intermediately predicts visual affordances and action primitives through direct supervision which provide the aforementioned auxiliary supervised guidance. We demonstrate that such joint learning and supervised guidance facilitates hierarchical task decomposition, assisting the agent to learn faster, achieve better driving performance and increases transparency of the otherwise black-box end-to-end network. We run our experiments to validate the MT-LfD framework in CARLA, an open-source urban driving simulator. We introduce multiple non-player agents in CARLA and induce temporal noise in them for realistic stochasticity. 
 
\end{abstract}

\section{Introduction}
A lot of recent work in autonomous driving has been focused on designing end-to-end learning networks for the task of driving using input images \citep{bojarski2016end, xu2016end, bojarski2017explaining, codevilla2017end, DBLP:journals/corr/abs-1709-05581}. Most works model this as an end-to-end regression problem of arriving at the control values using the input pixel values directly. Clearly, humans utilize a much richer hierarchical task decomposition pipeline while carrying out visuomotor tasks like driving. Rather than arriving at the exact throttle, brake and steering percentages directly from high-dimensional pixel values, our brains first decompose the image into recognizable objects and their intuitive states in the scene \citep{felleman1991distributed, kandel2000principles}, then use visual abstractions along with past experience to come up with approximately optimal high level decisions or action primitives for the given state, and lastly use these action primitives to arrive at the exact motor commands to be executed \citep{Bertenthal1996OriginsAE, Mataric00sensory-motorprimitives, flash2005motor, hart2010neural}. Such kind of hierarchical structure allows us to tackle hard problems like driving by decomposing them into multiple sub-tasks allowing for better generalization.

Studies in Neuroscience suggest that humans use a combintion of model-based and model-free techniques for sequential decision making \citep{glascher2010states, evans2008dual, fermin2010evidence, beierholm2011separate}. Most of the current end-to-end driving networks only utilize a model-free architecture, without explicitly trying to infer the state or model of the environment which can be used for planning. A combination of model-based and model-free approaches could greatly improve decision-making in autonmous driving settings.

Another challenge with such end-to-end learning systems is their black-box nature \citep{DBLP:journals/corr/abs-1801-00631, DBLP:journals/corr/abs-1708-08296, DBLP:journals/corr/RibeiroSG16}. Deep Learning techniques, in general, are highly opaque function approximators having millions of self-learned parameters, each of which might not have any human-interpretable significance. This is a huge setback in life-critical decision-making tasks like autonomous driving, where it is essential to not only arrive at the correct decision but also know how the decision was made by the system.  

In this work, we attempt to overcome these obstacles by proposing supervised auxiliary tasks that are learned by the network along with the main task of driving. We propose a set of `visual affordances' and `action primitives', that are annotated and used as auxiliary supervised tasks. The visual affordances are used to form abstract description of the visual scene in front of the vehicle and the action primitives provide an abstract description of the possible high-level actions required for driving. 

The auxiliary tasks serve a three-fold purpose. Firstly, the auxiliary tasks allow us to infuse a rich prior in the form of human knowledge into the system that assists the  final prediction task instead of expecting the network to learn all relevant knowledge from scratch. While trying to demonstrate the driving task, a human identifies the visual affordances and action primitives that are essential for driving, and provides the network with the auxiliary task of predicting these abstractions, thus assisting its decision-making with human knowledge. Secondly, the joint learning of the auxiliary task along with the main task of driving provides auxiliary guided supervision by forcing the network to predict intermediate representations like distance to vehicles and orientation with respect to the lane among others, that can be crucial to arrive at the final driving decision. This guided supervision allows the network to learn superior internal features thus allowing it to learn faster and generalize better. The auxiliary visual affordances can also be seen as an abstract description of the agent's environment and the joint-learning technique as a method of efficiently combining model-based and model-free techniques. Lastly, predicting the visual affordances and action primitives allows for better transparency in the network's learned internal representations, helping us better understand its decision-making process.  

We demonstrate our hypothesis in the CARLA simulator \citep{dosovitskiy2017carla}. All trials are started with random initialization of the player and non-player positions and goals and the non-player vehicles are induced with temporal noise to increase stochasticity of the multi-player dynamics. Such a stochastic setting makes it more difficult for the agent to overfit to a deterministic non-player policy and requires that the player have temporal information about the non-player agents to infer their present state without which it cannot derive it's own policy. The demonstrated expert policy is also extremely opportunistic as the demonstrator tries to not only avoid all collisions but also reach the goal locations in the least amount of time, generously overtaking vehicles if required. These changes make the driving conditions more realistic and challenging for a machine learning algorithm. 

\section{Related Work}
Learning from Demonstration or Imitation Learning \citep{schaal1997learning, argall2009survey, atkeson1997robot} and Reinforcement Learning \citep{sutton1998reinforcement, kaelbling1996reinforcement} have proven to be the key techniques for learning sequential decision-making tasks through demonstration and experience respectively. Reinforcement Learning requires a hand-crafted reward signal which the agent tries to maximize over time and in turn learns the desired policy through trial-and-error. The agent has to learn to assign credit to past actions faithfully in case of delayed rewards and also balance exploration and exploitation effectively to arrive at an ideal policy, both of which are extremely challenging \citep{sutton1984temporal, sutton1992introduction}. In Learning from Demonstration, on the other hand, the agent learns by directly trying to mimic the decisions of an expert demonstrator. Though originally designed for low-dimensional state-space tasks, with the advent of superior supervised function approximators \citep{krizhevsky2012imagenet, he2015delving, szegedy2017inception}, a lot of progress has been made in scaling these algorithms to very high dimensional state-space tasks enabling agents to learn policies directly from images \citep{DBLP:journals/corr/TaiL16a, mnih2015human, ratliff2009learning, bojarski2016end}. 

In this work, we use Learning from Demonstration framework to train an agent to learn the task of driving using high-dimensional observations in form of images. Learning from Demonstration along with deep function approximators have been used to tackle a lot of problems in robotics like indoor mobile robot navigation \citep{tai2016deep},  quad-rotor control in forest trials \citep{giusti2016machine}, robot-arm manipulation \citep{DBLP:journals/corr/DuanASHSSAZ17, DBLP:journals/corr/abs-1709-04905, DBLP:journals/corr/abs-1802-01557} among others. The closest to our work are the works of \cite{bojarski2016end} who show autonomous lane following using a single trained network, \cite{codevilla2017end} who demonstrate autonomous driving in CARLA using an additional conditional input from a high-level planner, \cite{DBLP:journals/corr/abs-1711-06459} who compare various contemporary networks for autonomous driving tasks and \cite{chowdhuri2017multi} who demonstrate multi-task and multi-modal behavior for autonomous driving. 

Multi-task learning (MTL) research shows the joint training of auxiliary related side-tasks along with the main task enhances the training performance \citep{caruana1998multitask, DBLP:journals/corr/abs-1203-3536}. MTL in neural networks \citep{DBLP:journals/corr/Ruder17a} has been successfully demonstrated in many tasks previously including text-to-speech conversion \citep{seltzer2013multi}, natural language processing \citep{collobert2008unified}, speech processing \citep{deng2013new} and computer vision \citep{DBLP:journals/corr/Girshick15, zhang2016augmenting}. In the field of sequential decision making, \cite{DBLP:journals/corr/LampleC16} demonstrate MTL for 3D game playing, \cite{DBLP:journals/corr/MirowskiPVSBBDG16} and \cite{DBLP:journals/corr/JaderbergMCSLSK16} demonstrate MTL in 3D maze navigation task whereas \cite{chowdhuri2017multi} utilize the MTL framework for autonomous driving. Instead of employing future control outputs as auxiliary tasks as shown by \cite{chowdhuri2017multi}, in this work we employ action and visual abstractions to guide the driving behavior.

Supervised learning of visual affordances for autonomous driving was introduced by \cite{DBLP:journals/corr/ChenSKX15}, though they use the predicted affordances to plan using a set of fixed rules whereas our network uses visual affordances as auxiliary tasks for the main task of driving. Action primitives can be inferred as sub-policies for the desired task. Learning hierarchical policies via demonstration is an active area of research \citep{byrne1998learning, demiris2005motor, DBLP:journals/corr/abs-1803-00590, DBLP:journals/corr/abs-1803-01840} and research in developmental psychology has also found evidence of hierarchical task decomposition during imitation in young children \citep{whiten2006imitation}. Our work decomposes the main task of driving into sub-policies which are used as auxiliary supervision to derive the final control commands.

\section{Multi-task Learning from Demonstration (MT-LfD) Framework}
\subsection{Learning from Demonstration}
We first begin by detailing the framework used to train the agent through expert demonstrations for the task of driving autonomously along with auxiliary task guidance. Learning from Demonstration (LfD) involves training an agent to try and imitate an expert demonstrator. At each time step $i$, the expert demonstrator is provided with an observation $o_{i}$, and the demonstrator provides the ideal action $u_{i}$ for that particular observation. A dataset $D = \{o_{i}, u_{i}\} _{i = 1} ^ N$ comprising of multiple episodic sequential roll-outs of the demonstrations is curated. 

LfD works on the assumption that if the demonstrator could deduce the ideal actions $u_{i}$ from the provided observations $o_{i}$, and if the demonstrator uses a consistent policy to determine the ideal actions $u_{i}$, there must exist a constant mapping function $F$ which maps the correlation between actions and observations $u_{i} = F( o_{i}) \forall i \in [1,N]$. In such a scenario, an agent parameterized by $\theta$ can be trained to obtain a policy $\pi(u_{i}/o_{i}; \theta)$ which maps the observations to actions. If a sufficiently expressive function approximator is used to train the agent, the parameters $\theta$ can be tuned such that the learned policy $\pi$ is almost equivalent to the demonstrated mapping function $F$. This can be done by tuning the parameters using the following update rule:
\begin{equation}
\label{eq:1}
\theta = \arg\min_\theta \sum_{i=1}^N loss(\pi(o_i), u_i) 
\end{equation}
and since we are assuming that $u_{i} = F( o_{i}) \forall i \in [1,N]$ the above equation is reducing the loss:
\begin{equation}
loss(\pi(o_i), F( o_{i}))
\end{equation}
and hence trying to make $\pi\approx F$. 

If the collected dataset $D$ is diverse enough to cover a large support of the distribution, the learned policy $\pi(u/o; \theta)$ could generalize to new unseen observations $o_j$ which are similar to the demonstrated observations, and predict a faithful control output $u_j$. Even so, distribution-mismatch is a common problem in LfD, where the demonstrated distribution does not match the test-time distribution, leading to unexpected compounding errors. A lot of methods have been suggested to overcome the distribution-mismatch problem \citep{DBLP:journals/corr/abs-1011-0686, levine2013guided, bojarski2016end}. In this work, we employ the noise injection method suggested by \cite{DBLP:journals/corr/LaskeyLHLMFG17}, by inducing noise in the agent during demonstration (details of which are in Section \ref{subsection4.2}) to force it to visit novel states and improve the demonstration distribution.

\subsection{Auxiliary Task Supervision}
Input observations $o_i$ include high-dimensional images and thus the underlying state $s_i$ of the system has to be inferred from these images for model-based decision making. Here we form an abstract description of the state $s_i$ using the visual affordances $v_i$. These local visual statistics allow the agent to learn an abstract local model of it's environment. The control commands $u_i$ predicted by the agent can also be decomposed into sub-policies. We decompose the task by enabling the agent to predict action primitives $a_i$ which are abstract descriptions of the policy. The visual affordances and action primitives are used as auxiliary supervised sub-tasks that are predicted by the multi-task network. 

The dataset $D$ is augmented with the visual affordances $v_i$ and action primitives $a_i$ annotation and thus becomes $D = \{o_{i}, u_{i}, v_i, a_i\} _{i = 1} ^ N$. Let $\pi(u_i/o_i')$, $\phi(v_i/o_i)$ and $\psi(a_i/o_i)$, jointly parameterized by $\theta$ denote the learnable functions to predict control, visual affordances and action primitives respectively. Since the policy $\pi$ is now guided by the predicted visual affordances and action primitives, policy $\pi$ is conditioned on $\phi$ and $\psi$ along with observation $o_i$ as $\pi(u_i/o_i, \phi(o_i), \psi(o_i))$ which we denote as $\pi(u_i/o_i')$. We can augment the update rule defined in Eq. \ref{eq:1} as follows:
\begin{equation}
\label{eq:3}
\theta = \arg\min_\theta \sum_{i=1}^N  loss(\pi(o_i'), u_i) + \alpha \, loss(\phi(o_i), v_i) + \beta \, loss(\psi(o_i), a_i) + \gamma (\theta)^2
\end{equation}
where $(\theta)^2$ is the L2 regularization loss and $\alpha, \beta, \gamma$ are hyperparameter coefficients for the auxiliary losses. As the learning progresses and the network is able to learn to faithfully predict visual affordances and action abstractions, $\phi(o_i) \approx v_i$ and $\psi(o_i) \approx a_i$. Thus eventually, the policy is conditioned on the visual affordances and action primitives as $\pi(u_i/o_i, v_i, a_i)$ and is guided by them.

\subsection{Network Architecture}
\label{sec:3.3}
\begin{figure}[h]
\includegraphics[scale=0.28]{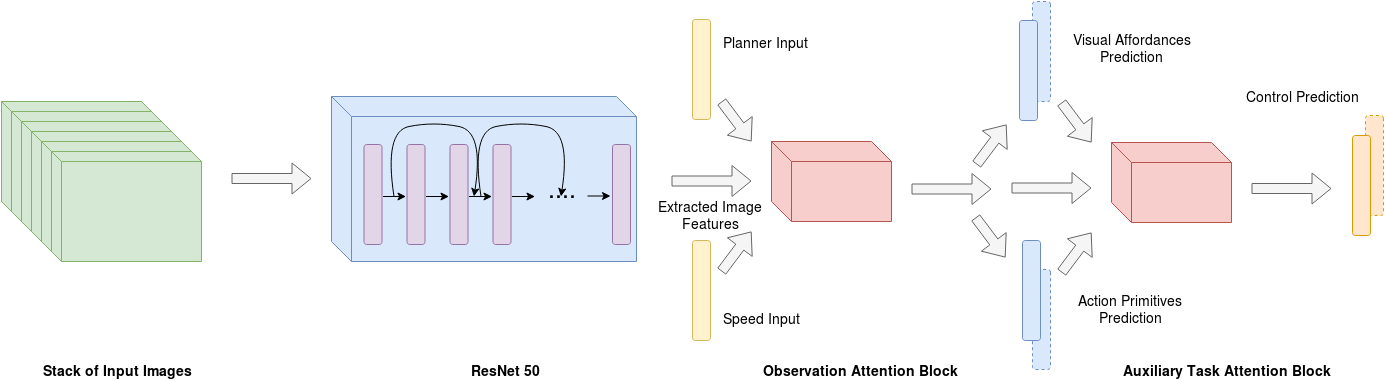}
\centering
\caption{High-level representational block diagram of the proposed network. }
\label{fig:image00}
\end{figure}

In this sub-section we outline the network architecture we use to perform MT-LfD. In our framework, the observations $o_i$ consist of stacked input images, forward speed of the player agent and a planner input which provides the high-level directions at intersections. To encode temporal information which is vital for stochastic multi-agent urban scenarios that we experiment with, we provide a history sequence of five images stacked along with the the current image at each time-step. The control predictions $u_i$ include the brake, throttle and steering percentage. Details about the visual affordances and actions primitives and their ground truth annotation scheme is presented in Section \ref{Sec:4.3}.

Figure \ref{fig:image00} shows a high-level representation of our proposed network. We use Resnet-50 \citep{DBLP:journals/corr/HeZRS15} to extract useful features from the stack of input images. Instead of adding or concatenating the speed and planner inputs we use a learnable soft-attention mechanism \citep{DBLP:journals/corr/BahdanauCB14} to attend to the extracted image features using the speed and planner inputs as represented by the input attention block in Figure \ref{fig:image00}. A learnable soft-attention mechanism allows for the low dimensional speed and planner inputs to have a large impact on the relatively higher-dimensional image feature vector. This attention mechanism is represented by a neural network layer and is jointly learned with the main network. The speed and planner inputs separately attend to the extracted feature vector and a union of the two attended features is taken to achieve a jointly attended vector. This attended observation feature is used to predict the visual affordances and action primitives. Lastly, in the Auxiliary Task Attention Block, a similar aforementioned double-attention mechanism is used to attend over the observation feature vector using the visual affordances and action primitives, which is then used to predict the final control predictions. The dashed-line vectors in Figure \ref{fig:image00} represent the ground-truth positions from where the gradients are jointly back-propagated.

% - oi consists of stacked high-dimensional images, forward speed measurement and the high level planner input\\
% - ui consists of the throttle, brake and steering values.\\
% - ResNet-50 is used to extract useful features from the stack of images.\\
% - the measurements - speed and planner input are then used to attend over the extracted features using a learn-able soft-attention mechanism.\\
% - a union over the attended features is taken and used to predict the visual affordances and action primitives.\\
% - the visual affordances and action primitives are then used to attend to the observation features using a learnable soft-attention.\\
% - a union of the attended observation features from visual affordances and action abstraction is taken and is passed through fully connected layers to predict the driving commands.  

\section{Experimental Setup}
\label{sec:num4}
\subsection{Environment Setup}
\begin{figure}[h]
\includegraphics[scale=0.25]{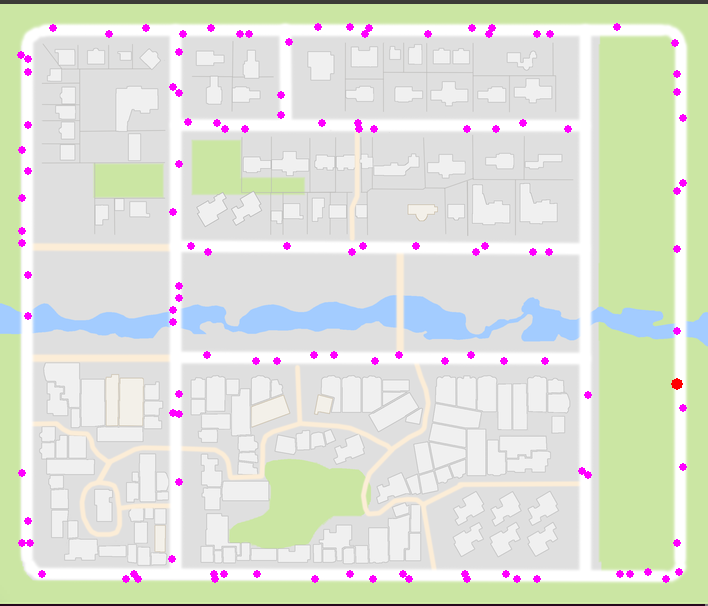}
\centering
\caption{Map of the town used for collecting the demonstrations. Pink dots indicate the initialized positions of the non-player agent vehicles where as the red dot indicates the initialized position of the player agent.}
\label{fig:image01}
\end{figure}

We use CARLA, an open-source 3D urban driving simulator, for our experiments. The use of CARLA allows for high-fidelity graphics and physics simulations of urban driving environment including diverse models of vehicles, pedestrians, houses, static obstacles, side-walks and intersections. For training, we collect data from a town with 2.9 km drivable urban roads. The map of the town is as shown in Figure \ref{fig:image01}.

To enable rich multi-agent behaviour, we initialize the map with 120 non-player vehicles and 140 non-player pedestrians during demonstration. The player as well as non-player agents are initialized at random start positions at the beginning of each episode to increase stochasticity of the system dynamics. The non-player pedestrians have a rich AI which enables diverse realistic behaviours. Each pedestrian is provided a goal location at random and  assigned a random maximum walking speed. The pedestrians try to reach the random goal point, without colliding with other static or dynamic obstacles and use a weighted navigation mesh to decide when to walk on the footpath or when and where to cross the road at a stochastically sampled angle to the road. The non-player vehicles are initialized with random model and random colours at the beginning of each episode and have a rich intelligent behaviour as well. Each non-player vehicle drives within lane, stops at traffic lights, follows the speed limit of the particular road it is travelling on, randomly samples turns at intersections and actively avoids all other static and dynamic obstacles.

\begin{figure}[h]
 
\begin{subfigure}{0.5\textwidth}
\includegraphics[scale=0.6]{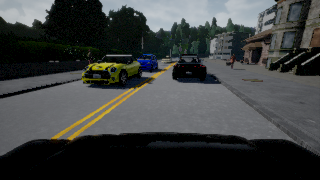} 
\caption{ }
\label{fig:subim1}
\end{subfigure}
\begin{subfigure}{0.5\textwidth}
\includegraphics[scale=0.6]{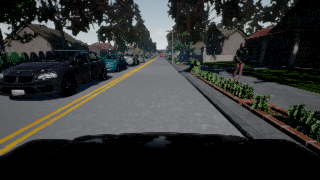}
\caption{ }
\label{fig:subim2}
\end{subfigure}

\begin{subfigure}{0.5\textwidth}
\includegraphics[scale=0.6]{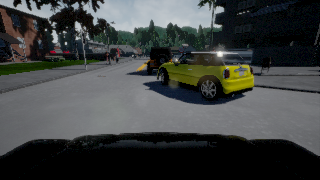}
\caption{ }
\label{fig:subim3}
\end{subfigure}
\begin{subfigure}{0.5\textwidth}
\includegraphics[scale=0.6]{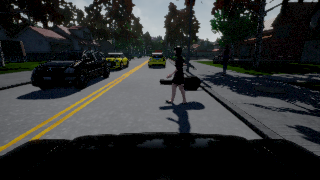}
\caption{ }
\label{fig:subim4}
\end{subfigure}

\begin{subfigure}{0.5\textwidth}
\includegraphics[scale=0.6]{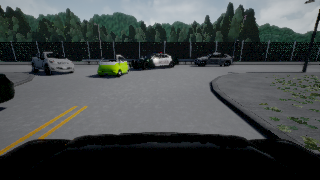}
\caption{ }
\label{fig:subim5}
\end{subfigure}
\begin{subfigure}{0.5\textwidth}
\includegraphics[scale=0.6]{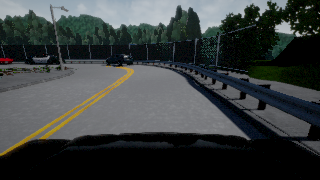}
\caption{ }
\label{fig:subim6}
\end{subfigure}
 
\caption{Sample simulation images as recorded by the front-facing RGB camera showing the diverse dynamics of the urban environment. \subref{fig:subim1} shows the player agent following a non-player agent vehicle. \subref{fig:subim2} shows a long queue of vehicles in the opposite lane. \subref{fig:subim3} shows a yellow vehicle cutting into the driving lane from a turn in front of the player agent. \subref{fig:subim4} shows a pedestrian carrying a guitar crossing the road in front of the player agent. \subref{fig:subim5} shows a busy  intersection with agents approaching from multiple directions. \subref{fig:subim6} shows an approaching left turn. }
\label{fig:image2}
\end{figure}

Even though the non-player vehicles have rich intelligent behaviour, their deterministic nature makes them highly unrealistic. A player agent could easily map such deterministic behaviour and overfit it's own behaviour. In such cases a player agent could easily derive it's own policy given a single current time-step image. To increase the stochasticity of the non-player vehicles, we induce temporal noise in the non-player vehicles. The temporal noise is enabled with some random probability and causes the non-payer vehicles to brake immediately and stop for a random duration sampled from a uniform distribution. This simple noise injection, prevents the player agent from overfitting to non-player vehicles policy and gives rise to realistic high-level urban traffic behaviours like erratic stopping, slow moving vehicular queues, busy multi-directional traffic at intersections, increased probability of overtaking maneuvers among others as can be seen in Figure \ref{fig:image2}.
% -map of town\\
% -number of pedestrians and vehicles\\
% -non-player vehicle AI\\
% -pedestrain AI\\
% -noise in non-player agents\\
% -emergent traffic behavior due to noise\\

\subsection{Data Collection}
\label{subsection4.2}
In our framework, a human driver is an expert demonstrator and provides the ground truth driving commands using CARLA that the network is trained to imitate. A front facing RBG mono-camera of $320 X 180$ pixel resolution and $100^{\circ}$ FOV is mounted on a simulated vehicle for collecting visual information while driving. The expert demonstrator looks at a similar first-person view of a higher resolution ($1280 X 720$) while driving the vehicle to collect data. The complications of record-mapping \citep{argall2009survey} are prevented by having the demonstrator use a similar visual view point for demonstration. 

\begin{figure}[h]
\includegraphics[scale=0.05, angle=-90]{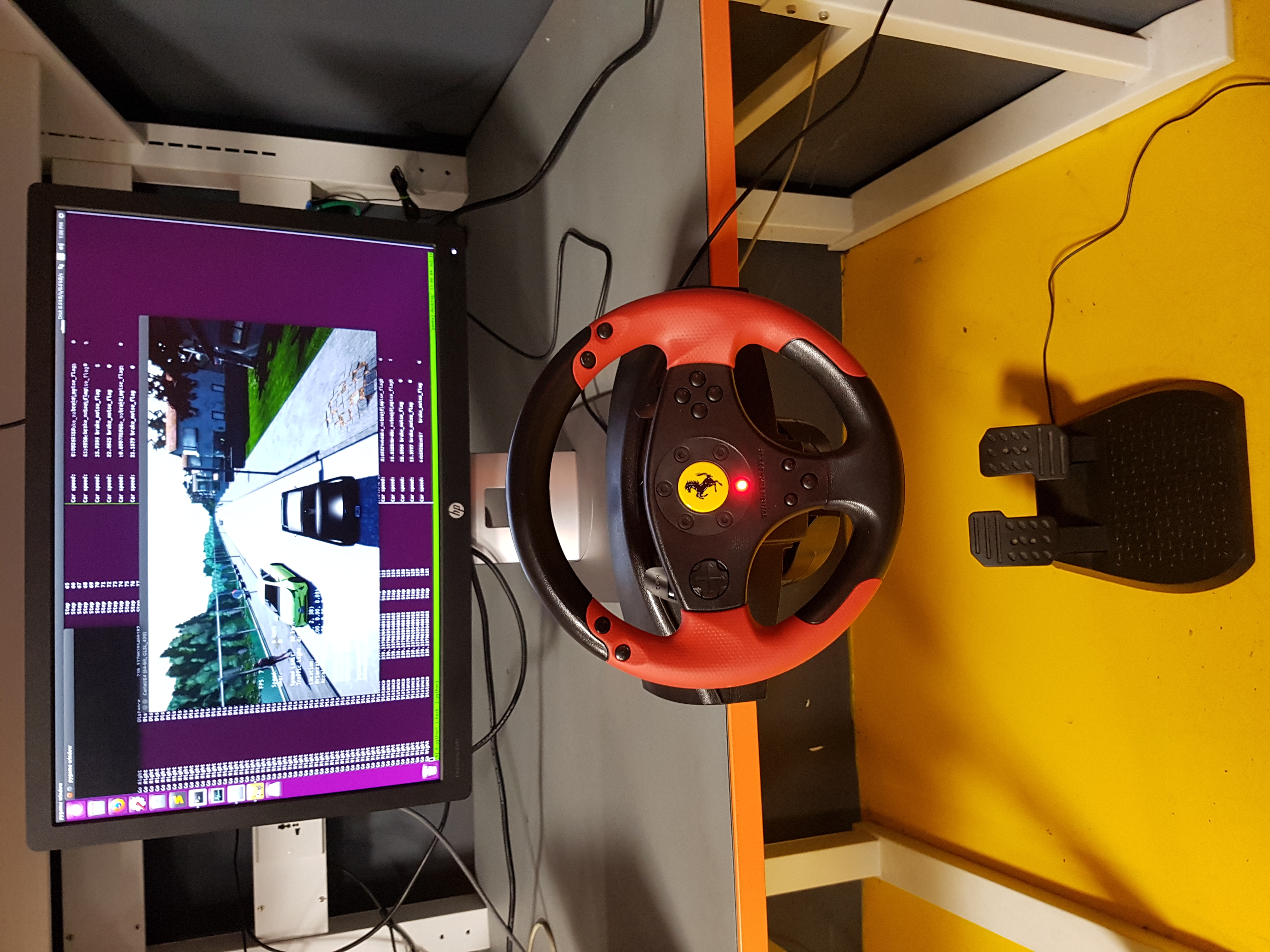}
\centering
\caption{Setup used by expert demonstrator to provide demonstrations in CARLA.}
\label{fig:setupimage}
\end{figure}

The demonstrator is equipped with the task of demonstrating the ideal brake, throttle and steering commands for each observation frame. The demonstrator does this by using a gaming steering wheel and throttle and brake pedals, which allow for easier demonstrations and enable analog inputs into the system. The setup we use for demonstration is shown in Figure \ref{fig:setupimage} At the beginning of each episode, a collection of random goal points are selected and sequentially provided to an A* planner which uses the town map to plan the shortest route to the goal locations. The A* planner provides the demonstrator with one of the four high level commands (go left, go right, go straight, follow lane) which the demonstrator follows while providing demonstrations. The demonstrator avoids all static and dynamic obstacles and also tries to stay on lane as much as possible. The demonstrator also tries to reach the goal in the least amount of time, even overtaking long queues of vehicles if required, while trying to drive within the above mentioned constraints. This results in an extremely rich demonstration policy making it more realistic and difficult to imitate in novel scenarios.    
% Demonstration:\\
% - brake, throttle and acceleration control commands demonstrated using gaming steering wheel and pedals.\\
% - driver tries to follow high-level A-star planner which provides directions to random goal points.\\
% - driver tries to avoid static and dynamic obstacles.\\
% -opportunistic driving behaviour while demonstration. overtaking vehicles when possible\\
% - higher resolution first-person view while driving the vehicle to collect data.\\

To overcome the distribution mismatch problem, we induce noise in the player agent and record the demonstrations of the agent recovering from the noise as provided by the expert demonstrator. Unlike \cite{codevilla2017end} who induce a fixed duration temporal noise in the steering, we induce both positive and negative temporal noises in the steering, brake and throttle; the probability and duration of which are sampled from a uniform distribution.  

We collect $150,000$ frames of images at about 6 fps resulting from approximately 7 hours of driving demonstration spread across 82 episodes. The player agent data collected include RGB images, high-level planner command and speed provided by CARLA which form our input observation $o_i$, and steering, brake and throttle demonstrations provided by the expert demonstrator which forms the action $u_i$. Other auxiliary player and non-player measurements are also collected which help us in annotating the ground-truth visual affordances $v_i$ and action primitives $a_i$ as discussed below.   
\footnotesize
% \begin{table}[!h]
% \label{table:1}
% %\hspace{20}
% \begin{tabular}{l}
% \hline
% \multicolumn{2}{c}{\textbf{ Visual Affordances \qqad}} \\
% \hline
% Distance to vehicle ahead in same lane\\
% Velocity of vehicle ahead in same lane\\
% Distance to vehicle ahead in opposite lane\\
% Velocity of vehicle ahead in opposite lane\\
% Percentage player in opposite lane\\
% Percentage player on sidewalk\\
% Distance to approaching intersection\\
% Distance to approaching left turn\\
% Distance to approaching right turn\\
% Longitudinal distance to pedestrian approaching\\
% Lateral distance to pedestrian approaching\\
% Longitudinal distance to pedestrian departing\\
% Lateral distance to pedestrian departing\\
% \end{tabular}
% \hspace{50}
% \begin{tabular}{l}
% \hline
% \multicolumn{1}{c}{\textbf{Action Primitives}} \\
% \hline
% Do nothing\\
% Speed up\\
% Slow down\\
% Halt\\
% Turn left\\
% Turn right\\
% Cut out of the lane\\
% Cut into the lane\\
% \\
% \\
% \\
% \\
% \\
% \end{tabular}\\
% \caption{List of visual affordances and action primitives used for auxiliary supervision}
% \label{table:1}
% \end{table}

\begin{table}[!htb]
    \begin{minipage}{.5\linewidth}
      \centering
        \begin{tabular}{l}
        \hline
        \multicolumn{1}{c}{\textbf{ Visual Affordances}} \\
        \hline
        Distance to vehicle ahead in same lane\\
        Velocity of vehicle ahead in same lane\\
        Distance to vehicle ahead in opposite lane\\
        Velocity of vehicle ahead in opposite lane\\
        Percentage player in opposite lane\\
        Percentage player on sidewalk\\
        Distance to approaching intersection\\
        Distance to approaching left turn\\
        Distance to approaching right turn\\
        Longitudinal distance to pedestrian approaching\\
        Lateral distance to pedestrian approaching\\
        Longitudinal distance to pedestrian departing\\
        Lateral distance to pedestrian departing\\
        \end{tabular}
    \end{minipage}%
    \begin{minipage}{.5\linewidth}
      \centering
        \begin{tabular}{l}
        \hline
        \multicolumn{1}{c}{\textbf{Action Primitives}} \\
        \hline
        Do nothing\\
        Speed up\\
        Slow down\\
        Halt\\
        Turn left\\
        Turn right\\
        Cut out of the lane\\
        Cut into the lane\\
        \\
        \\
        \\
        \\
        \\
        \end{tabular}\\
    \end{minipage} 
    \caption{List of visual affordances and action primitives used for auxiliary supervision}
    \label{table:1}
\end{table}

\normalsize
% -sequence of frames collected of resolution.. at fps.. \\
% -player measurements collected and used for training. non player measurements collected and used for finding auxiliary annotations.\\
% -episodic sequences collected till demonstrator crashes or has no passage due to traffic.\\
% -amount of data collected.\\
\subsection{Auxiliary Tasks and Data Annotation}
\label{Sec:4.3}
In this subsection, we describe the visual affordances and action primitives used for auxiliary supervision. We identify 13 different visual affordances that are critical to describe the local state $s_i$ of the system and 8 different action primitives that decompose the policy effectively. A list of all the visual affordances and action primitives is provided in Table \ref{table:1}. A few visual affordance measurements like 'Percentage player in opposite lane' and 'Percentage player on sidewalk' are directly provided by CARLA, but most others have to be derived using a fixed set of rule-based mechanism. Multiple player and non-player measurements are collected which are not directly used for training the network, but are used to derive the ground truth annotations for the visual affordances. These measurements include player global position, player yaw, non-player agent positions, non-player agent yaw and non-player agent velocity. Action Primitives are derived using fixed rules to classify the demonstrated steering, brake and throttle control values.    

% - 13 visual affordances namely..\\
% - automatic annotation rules for visual affordances..\\
% - 8 action primitives namely.. \\
% - automatic annotation rules for action primitives.. \\

% \footnotesize
% \begin{center}
%     \begin{tabular}{l    r}
    
%     \begin{left}
% \begin{tabular}{ |l| } 
% \hline
% \multicolumn{3}{|c|}{\textbf{Visual Affordances}} \\
% \hline
% \multirow

% Distance to vehicle ahead in same lane\\
% \hline Velocity of vehicle ahead in same lane\\
% \hline Distance to vehicle ahead in opposite lane\\
% \hline Velocity of vehicle ahead in opposite lane\\
% \hline Percentage player in opposite lane\\
% \hline Percentage player on sidewalk\\
% \hline Distance to approaching intersection\\
% \hline Distance to approaching left turn\\
% \hline Distance to approaching right turn\\
% \hline Longitudinal distance to pedestrian approaching\\
% \hline Lateral distance to pedestrian approaching\\
% \hline Longitudinal distance to pedestrian departing\\
% \hline Lateral distance to pedestrian departing\\
% \hline
% \end{tabular}
% \end{left}

% \begin{right}
% \begin{tabular}{ |l| } 
% \hline
% \multicolumn{3}{|c|}{\textbf{Action Primitives}} \\
% \hline
% \multirow
% \hline Do nothing\\
% \hline Speed up\\
% \hline Slow down\\
% \hline Halt\\
% \hline Turn left\\
% \hline Turn right\\
% \hline Cut out\\
% \hline Cut in\\
% \hline
% \end{tabular}
% \end{right}

%     \end{tabular}
% \end{center}

\section{Results}

We train the proposed network in Section \ref{sec:3.3} using the expert demonstrated data for the control values and the automatic rule-based ground truth generated for the visual affordances and action primitives. Out of the 150000 images collected, a sequence of 10000 images was held out for validation and the rest was used for training. The control prediction and visual affordances prediction are regression problems and thus we use Mean Squared Error loss for $\pi(o_i')$ and $\phi(o_i)$ whereas the action primitives prediction is a mutually non-exclusive multi-class classification problem and thus we use a multi-class cross entropy loss for $\psi(o_i)$.  Data augmentation including Gaussian blurring, Gaussian noise, pixel dropouts, gray-scaling, contrast change and intensity shits were used to help with better generalization. We use Adam optimizer with an initial learning rate of $3e^{-4}$ and use learning rate decay. We set $\alpha$, $\beta$ and $\gamma$ to $0.1$, $0.2$ and $1e^{-5}$ respectively. The network is trained for approximately 15 hours on a GPU. We call our proposed network as \textit{MT-LfD} through the rest of this section.

\begin{figure}[h!]
\begin{subfigure}{0.5\textwidth}
\includegraphics[scale=0.50]{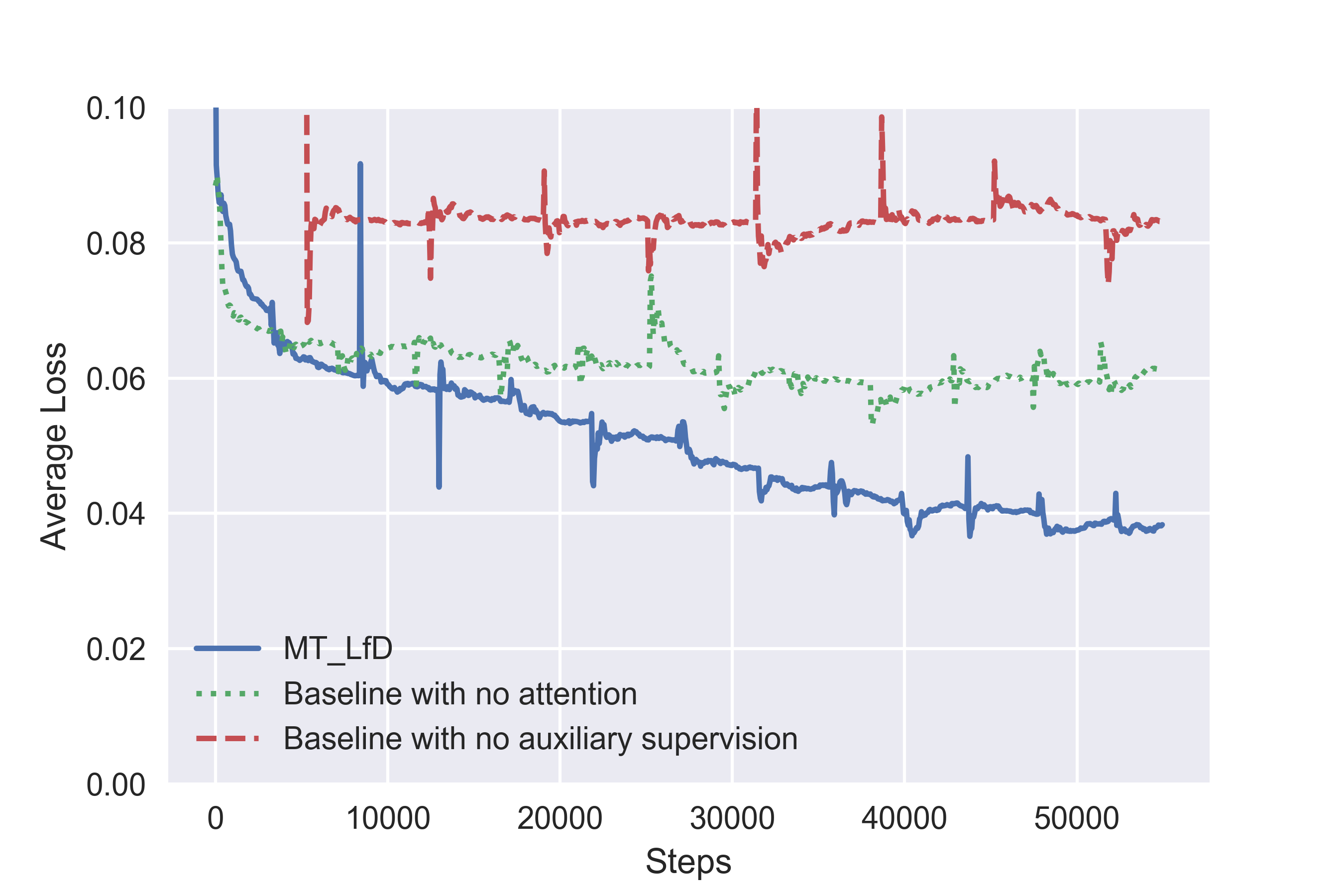}
\caption{ }
\label{fig3:subim1}
\end{subfigure}
\begin{subfigure}{0.5\textwidth}
\includegraphics[scale=0.50]{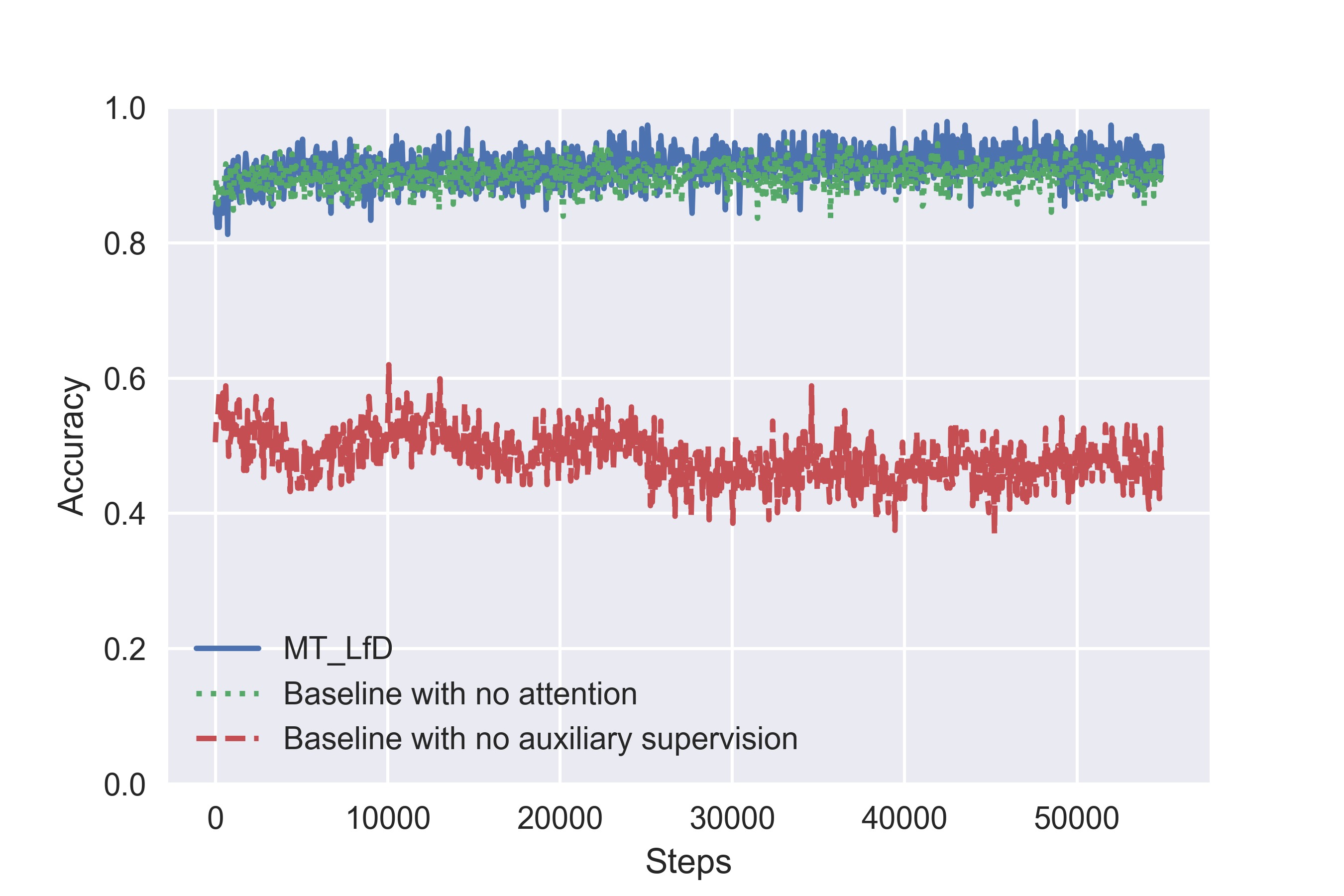}
\caption{}
\label{fig3:subim2}
\end{subfigure}
\caption{\subref{fig3:subim1} Average MSE loss for the control prediction task on the validation set as the training progresses (Lower is better). \subref{fig3:subim2} Accuracy for the multi-class classification task of predicting the action primitives on the validation set as the training progresses (Higher is better).}
\label{fig:image3}
\end{figure}

Figure \ref{fig:image3} shows the plots of our results. We compare the results of \textit{MT-LfD} with 2 baselines to verify the performance of \textit{MT-LfD}. The first baseline which we call \textit{Baseline with no attention} is an ablated version of the \textit{MT-LfD} network architecture in which we use simple concatenation instead of learnable soft-attention to combine the auxiliary information along with the visual information. The second baseline which we call \textit{Baseline with no auxiliary guidance} has the same architecture as \textit{MT-LfD} with $\alpha$ and $\beta$ set to zero. This essentially prevents the loss from the visual affordances and action primitives from back propagating through the network and thus removes the auxiliary guidance provided by their predictions. Figure \ref{fig3:subim2} shows the action primitives prediction accuracy on the held-out validation set. Since \textit{Baseline with no auxiliary guidance} does not have a loss for action primitives, it can be observed that it performs almost on par with random sampling whereas the others perform much better at the classification task.

  We keep the rest of the hyperparameters same as \textit{MT-LfD} for the baselines. Figure \ref{fig3:subim1} shows the mean-squared error loss for the final control prediction task on the held-out validation set. \textit{MT-LfD} has lower loss as compared to \textit{Baseline with no attention} thus validating our use of the soft-attention mechanism. \textit{MT-LfD} also has a lower loss and converges much faster than \textit{Baseline with no auxiliary guidance} thus proving the necessity of the guided auxiliary supervision provided by the visual affordances and action primitives, validating our hypothesis. 

\section{Summary}
We demonstrate Multi-task Learning from Demonstration for end-to-end learning of autonomous driving, jointly supervised and guided by visual affordances and action primitives. We present our network architecture which uses ResNet-50 and learnable soft-attention mechanisms to combine the auxiliary task predictions with the observation information for the final driving task prediction. We show that our proposed MT-LfD framework outperforms vanilla LfD and MT-LfD without attention in the main task of vehicle control prediction. We thus, validate our hypothesis that the joint learning of the auxiliary tasks and the employing their predictions to guide the final control prediction is able to enhance the speed and the performance of learning. 

Nevertheless, our network does not outperform previous work in more deterministic setups with controlled number of agents and maximum speed. Driving in a realistic scenario with a highly stochastic multi-agent setup and realisitic driving demonstrations still remains a challenging open problem. It remains to ben seen how memory-augmented networks perform in such a scenario. A prospective future work could be to use Inverse-Reinforcement Learning to derive the intention of the demonstrator and use model based-statistics along with such inferred intention to derive an ideal driving policy. Another orthogonal prospective work could be to train a network to plan on the model-based statistics and use the planning network along with the model-free prediction network to derive the driving policies. More work also needs to be done in making driving simulators more realistic to human-driving scenarios for autonomous driving experiments. 

\bibliographystyle{plainnat} % or try abbrvnat or unsrtnat
\bibliography{bibliography} 

\end{document}